\documentclass[a4paper]{article}
\usepackage{times}
\usepackage{latexsym}
\usepackage{amssymb}
\usepackage[dvipsnames]{xcolor}

\usepackage[T1]{fontenc}

\usepackage[utf8]{inputenc}

\usepackage[symbol]{footmisc}

\usepackage{hyperref}
\usepackage{url}
\usepackage{booktabs}
\usepackage{multirow}
\usepackage{comment}
\usepackage{graphicx}

\usepackage{INTERSPEECH2022}

\title{Data Augmentation for Dementia Detection in Spoken Language}
\name{Anna Hlédiková$^1$†\thanks{†Authors contributed equally.}, Dominika Woszczyk$^1$†$^*$\thanks{$^*$Work done outside of Amazon.}, Alican Akman$^1$, Soteris Demetriou$^1$, Björn Schuller$^{1,2}$}

\address{$^1$Department of Computing, Imperial College London, London, UK\\
$^2$Chair of Embedded Intelligence for Health Care and Wellbeing, University of Augsburg, Germany}
\email{anna.hledikova20@imperial.ac.uk, d.woszczyk19@imperial.ac.uk, a.akman21@imperial.ac.uk, s.demetriou@imperial.ac.uk, bjoern.schuller@imperial.ac.uk}

\begin{document}

\maketitle

\begin{abstract}
Dementia is a growing problem as our society ages, and detection methods are often invasive and expensive. Recent deep-learning techniques can offer a faster diagnosis and have shown promising results. However, they require large amounts of labelled data which is not easily available for the task of dementia detection. One effective solution to sparse data problems is data augmentation, though the exact methods need to be selected carefully. To date, there has been no empirical study of data augmentation on Alzheimer's disease (AD) datasets for NLP and speech processing. In this work, we investigate data augmentation techniques for the task of AD detection and perform an empirical evaluation of the different approaches on two kinds of models for both the text and audio domains. We use a transformer-based model for both domains, and SVM and Random Forest models for the text and audio domains, respectively. We generate additional samples using traditional as well as deep learning based methods and show that data augmentation improves performance for both the text- and audio-based models and that such results are comparable to state-of-the-art results on the popular ADReSS set, with carefully crafted architectures and features\footnote{The code is available at\: https://github.com/hl-anna/DA4AD}.
\end{abstract}

\section{Introduction}
\label{sec:intro}

Dementia is the common term to describe a decline of cognitive abilities, such as memory, problem-solving or language that can severely impact an individual's communication abilities and everyday life in general. This condition, most often caused by Alzheimer's disease (AD), affects close to 60 million people and is the 7th leading cause of death, globally \cite{AlzWHO}.

Fortunately, there has been a great focus across different fields not only to develop treatment for the condition but also to detect it, as an early diagnosis is key to help individuals to take control, plan their future, and delay the symptoms.

Recent challenges such as ADReSS and ADReSSo brought the focus to using speech and/or transcripts for AD detection \cite{luz2020alzheimer, LuzEtAl21ADReSSo}. Detecting dementia, especially in its early stages, is a complex task which requires identifying subtle changes to semantics, vocabulary, or sentence-level structure. Information from audio data can provide beneficial information, e.\,g., about pauses or speech rate \cite{Yuan20}. Subsequent work explored large pre-trained models and various DNN embeddings to extract fine-grained representations for text and audio features and achieved good results on the challenge held-out test set, considering the limited data available \cite{Zhu21, Balagopalan20, Cummins20, Yuan20,syed2021automated,glass2021classifying}. Linguistic approaches have shown to be the most discriminative, even with simple models such as SVM. However, current state-of-the-art models implement hand picked features and tailored feature engineering techniques \cite{martinc21temporal}. Audio classifiers benefit greatly from pre-trained embeddings such as the one introduced by \cite{gong2021ast}, which have the advantage of being automatically extracted, compared to designed feature sets such as ComParE or eGeMAPS~\cite{eyben2015geneva}.

Collecting quality data on AD, be it transcripts or audio samples, can be difficult. Patients are hard to access, have to be evaluated for their mental state and the sessions have to be properly recorded and transcribed. Machine learning with limited data is a common challenge across tasks and domains, as limited amount of training data hinders the models' ability to generalise well and train properly. Some approaches addressing this problem explore different architectures and ensemble models~\cite{iliasmultimodal,syed2021automated,santander2022semantic,an2020deep}, feature engineering ~\cite{garla2012ontology,haider2019assessment}, and data augmentation techniques~\cite{Roshanzamir21,guo2021crossing,ye2021development,liu2021detecting}. In this work, we focus on the latter. Data augmentation has been quite successful in image processing tasks, as one can perform many transformations without changing the object and label of the images. Although some of those techniques can be transferred to speech processing, they need to be carefully chosen for the task at hand and take into account the time dimension, too. Augmenting text data is more challenging, as it is highly sensitive to changes and one must be careful not to affect key label-specific characteristics. Nevertheless, several techniques have been explored by~\cite{feng2021survey} and~\cite{chen2021empirical} in surveys which cover various tasks related to NLP or speech processing, extensively. Given the multi-modal nature of AD detection, suitable augmentation strategies are required for both audio and text. Individual text augmentation techniques have been applied in AD detection work \cite{Roshanzamir21,liu2021detecting,ye2021development}, but no work offers an overview of augmentation strategies.

In this paper, we aim to expand the scope of previous surveys on data augmentation and evaluate label preserving data augmentation techniques on the task of Alzheimer's disease detection for both NLP and speech processing. We examine a total of 17 approaches (7 for text, 10 for audio) and evaluate them on baseline and SOTA models. We work with the two domains separately, as well as with a fusion of both. Our experiments show that the selected data augmentation methods were able to preserve the class labels and that augmentation itself is effective in improving the model's generalisation abilities on the ADReSS dataset. This has been shown both using neural-based and more traditional models, on both the text and audio domains.
We believe that this work can contribute to the field and offer an inexpensive way to improve AD detection methods in the future.
Our contributions are as follows:

\begin{comment}
\begin{enumerate}
\item We identify and categorise NLP and speech processing data augmentation techniques for the task of dementia detection.
\item We compare the methods empirically with limited labelled data and evaluate their level of label preservation.
\item We further analyse the dysfluency score of each data augmentation technique. 
\end{enumerate}
\end{comment}

\begin{enumerate}
\item We empirically evaluate text and speech data augmentation techniques for the task of dementia detection.
\item We analyse the level of label preservation for each augmentation. 
\item We achieve performance on par with state-of-the-art model for both text and audio domains via data augmentation.

\end{enumerate}

\section{Data Augmentation Strategies}
\label{sec:methods}

\subsection{Text Augmentations}

\subsubsection{Noise}

Noise can be introduced by means of random deletions, insertions, or substitution at character, word, or sentence level. For the text domain, we employ three noise injection methods: sentence deletion (SD), easy data augmentation (EDA)~\cite{wei2019eda}, and Mixup~\cite{zhang2017mixup}. For the SD approach, we delete one or more randomly selected sentences from each transcript. We found that removing one to four sentences (at random) yields the best results, so the average SD operation reduces a transcript by 10\,\%. The Mixup approach splits a transcript into two halves and swaps these with a different sample within the same class. EDA randomly replaces, inserts, or deletes words within a document. This approach can affect the semantics or introduce grammar mistakes. We follow the EDA implementation from the original paper and set $\alpha=0.05$.

\subsubsection{Lexical substitution}
An intuitive approach to augmenting data while preserving the label is to substitute words with their synonyms. A simple option is to replace words with their WordNet~\cite{wordnet1995} synonyms or hyperonyms from the Knowledge Graph. The words to substitute can be sampled at random, using TFIDF-weights or by analysing the rate of change when the word is deleted. However, this is a tricky operation in case of AD detection. Words might be substituted by alternatives causing the sample to loose its discriminative features, for example by going from familiar language to elevated vocabulary. Some works use pre-trained embeddings and look for close words given the feature space, using geometric distances. These approaches suffer from the same issues. To tackle this limitation, one can use contextual augmentation, i.\,e., use language models to replace words given a sentence and learnt priors for each label \cite{kobayashi2018contextual}. In this work, we implemented a version of contexutal augmentation using the NLPAug library \cite{ma2019nlpaug}, based on the RoBERTa model. We set $p=0.1$ and $top\_k=20$, i.\,e., replacing 10\,\% of words by picking randomly for the top 20 alternatives.

\subsubsection{Paraphrase}

Paraphrasing consists in rewording or changing the structure of a document while maintaining its original meaning. Earlier approaches include statistical and rule-based models or selecting sub-sentences from a pool of stored templates. More recently, neural models have been adopted to generate paraphrases directly in an end-to-end fashion. A sub-task of paraphrasing is back-translation (BT), which creates new samples by translating text from one language to another and back to the original one, thus exploiting differences in structure and vocabulary across languages. Similarly, text summarisation aims to restructure sentences in a concise way. This approach is less desirable diversity-wise, as the target space is more limited. We split each document on sentence level and run the pre-trained paraphrase model Pegasus~\cite{zhang2020pegasus}, a model with 223M parametres, to generate paraphrases of length $n=60$ tokens and with temperature $t=1$.

\subsubsection{Text Generation}

Generative models are a popular approach to data augmentation for text. Rule-based text generation methods have now been supplanted by DNNs, sequence-to-sequence, and transformer-based methods with large models trained on even larger datasets. One such model is OpenAI's GPT-2~\cite{radford2019language}, which we use to perform text generation. GPT-2 is pretrained on the WebText corpus -- a large collection of internet blogs, posts, and pages, but can be finetuned with a relatively small set of samples to adapt the feature distribution. We use Hugging'face's\footnote{https://huggingface.co/gpt2} model with 117M parametres. The model is finetuned on the entire training set, conditioned with the sample label class. The label is then used as a prompt for the model at inference time.

\subsection{Audio Augmentations}
\subsubsection{Standard transformations}
 Standard transformations can be defined as a function $g$ applied on the sample $x$ at time step $t$ such that $x'(t) = g(x(t))$. We apply these transformations via the \texttt{nlpaug}\footnote{https://github.com/makcedward/nlpaug} python package.\\
\textbf{Noise addition} is one of the simplest transformations. The noise $\epsilon$ is usually sampled from a Gaussian distribution $N(\mu,\sigma)$, in our case $N(0,1)*0.002$, and is applied at step $t$ either in the time or frequency domain. The new sample can be defined as $x'(t) = x(t) + \epsilon$.\\
\textbf{Time stretching} changes the speed of a signal by a factor of $\alpha$, without changing the pitch. For $\alpha < 1$, the signal is slowed down and lengthened and sped up for $\alpha > 1 $.\\
\textbf{Pitch shifting} perturbs the pitch by $n$ fractions of an octave. The signal is passed through a Short-Term Fourier Transform (STFT) to obtain the spectrogram in the frequency domain to apply the change to the pitch, without changing the length or speed of the signal. We adjust the pitch by a scale factor in $(-10,10)$.\\
\textbf{Time shift} adds padding on the left or the right of a signal. We shift time of the whole audio by a duration of $0.5$ seconds.\\
\textbf{Loudness shift} increases or decreases the magnitude of a signal $x$ by $\beta$ DB or by scaling the signal by a factor $\alpha$ s.\,t.\ $x'(t) = \alpha x(t)$, where the volume is decreased for $\alpha <1$.  We adjust loudness of each sample by a scale factor within $(0.3,3)$.\\
\textbf{Normalisation} of the signal can be performed given minimum and maximum amplitudes or by standardising it. We simply normalise the signal by dividing it by the maximum amplitude.\\
\textbf{Time masking} consists in masking consecutive time bands on a log Mel-spectrogram, where $t$ $\epsilon [t_0,t_0+t]$ is sampled from a uniform distribution $U(0,T)$, $t_0\sim U(0,T-t)$, and $T$ is the maximum length of masking steps. We apply masking operation with a coverage of $0.3$.\\
\textbf{Frequency masking} applies a mask on consecutive frequency bands on a log Mel-spectrogram, where $f\sim U(f_0,f_0+f)$ and $f_0 \sim [0,V-f]$, where $V$ is the maximum number of Mel-frequency channels. We implement two strategies that combine several transformations: \textit{Random} and \textit{SpecAugment}. The \textit{Random} strategy selects a transformation for each sample at random and \textit{SpecAugment} combines time warping, frequency, and time masking on the log Mel-spectrograms~\cite{park2019specaugment}. 
\subsubsection{Vocal Tract Length Perturbation (VTLP)}
Vocal Tract Length Perturbation (VTLP) operates on the log-Mel-spectrogram itself~\cite{jaitly2013vocal}. Similarly to time warping, VTLP performs warping on a frequency $f$.
\begin{comment}
such that
\begin{equation}
  f' = \left \{ \begin{array}{ll} 
  fw & f \leq F_{hi}^{\frac{min(w,1)}{w}} \\
    \frac{S}{2} - \frac{\frac{S}{2} - F_{hi}min(w,1)}{\frac{S}{2}-F_{hi}\frac{min(w,1)}{w}} & otherwise,
    \end{array}
    \right
\label{eq:vtlp}
\end{equation}

where $w$ is a warp factor sampled from a uniform distribution, $S$ is the sampling frequency and $F_{hi}$ is a boundary frequency chosen such that it covers the relevant
formants, in this case determined by human range and gender.
\end{comment} 
We apply a VTLP operation on the whole audio with a scaling factor of $(0.5,3)$.
\subsubsection{Generative Models}
Deep neural networks are now dominating generative models such as auto-encoders, generative adversarial networks (GANs) or sequence-to-sequence models such as LSTM networks. We perform zero-shot voice conversion using a FragmentVC~\cite{lin2021fragmentvc} pre-trained model\footnote{https://github.com/yistLin/FragmentVC}. It performs better at speaker identity conversion than popular state-of-the art AutoVC \cite{qian2019autovc} while preserving prosody and durations, which is actually desirable in our case to preserve class-specific style.

\section{Experiments \& Results}
\label{sec:eval}

\subsection{Datasets}
\subsubsection{ADReSS and Dementiabank} We work with two datasets, the ADReSS dataset~\cite{luz2020alzheimer}\footnote{The dataset is accessible upon request at http://www.homepages.ed.ac.uk/sluzfil/ADReSS/ and is governed by the Creative Commons CC BY-NC-SA 3.0 license.} and Dementiabank.\footnote{https://dementia.talkbank.org/ }.
Both consist of recordings of a Cookie theft picture description task from individuals diagnosed with various stages of dementia. The datasets also contain manual transcriptions of the samples with dysfluency annotations, and the ADReSS set is a subset of Dementiabank. The speakers in the ADReSS set are American and have been selected to balance for gender and age, and they split evenly into control (HC) and dementia (AD) groups (54 train and 24 test for each class, total of 156 samples). Dementiabank has three times as many samples, but is imbalanced, lacks a fixed test set and some samples contain a lot of noise. The main focus of this paper is thus on the ADReSS set, which is of better quality and allows for an easy comparison of results with other works. However, we also examine the effect of augmentation on a larger dataset such as Dementiabank.

\subsubsection{Preprocessing}
Transcripts from both datasets were annotated using the CHAT coding system which includes POS tags and grammatical dependencies but also pauses and other dysfluencies. We read each CHAT file using the python library \texttt{pylangacq}~\footnote{https://pylangacq.org/}. We retain the cleaned sentences uttered by the participant, with no additional dysfluency annotation. 
For the audio recordings, we use the normalised, noise reduced wavefiles sampled at 44\,kHz. We split them into chunks of $10s$ with a stride of $2s$ and apply the data augmentation either on the signal or spectrograms computed using the \texttt{librosa}~\footnote{https://librosa.org/doc/latest/index.html} Python library.

\subsection{Models}

\subsubsection{Text-based detection models} 

To perform the linguistic classification task, we reimplement~\cite{Yuan20, Balagopalan20} using a pre-trained BERT model on the transcripts. We use the base size model (12 layers) with the following patametres: learning rate of 2e-5, 8 epochs, and input length limit
of 256 tokens. We feed the whole document as a datapoint, keeping the dysfluencies annotations. To focus on text only, we ignore the pauses extracted for the audio samples. The BERT model is widely used in the field of NLP and the authors from~\cite{Yuan20} have achieved SOTA results on the ADReSS Challenge 2020. As a more traditional approach, we also implement an SVM with an RBF kernel with $gamma=0.01$  and $C=1$. We use TF-IDF features as input. 

\subsubsection{Acoustic-based detection models} 

The audio classifier we use in our experiments is the Audio Spectrogram Transformer (AST) \cite{gong2021ast}, which adapts the transformer architecture for audio. We extract Mel spectrograms as input to the AST model. We tune the model for 2 epochs, using a learning rate of 1e-6 and the Adam optimiser. Similarly as for text, we also implement a simpler baseline which is a Random Forest (RF) model. We extract eGeMAPS~\cite{eyben2015geneva} sets of features to train the RF model, which has been shown to be a better optimised set of features compared to ComParE on the DementiaBank dataset~\cite{luz2020alzheimer}. The RF model is trained with $100$ trees.

\subsubsection{Fusion} Our fusion model consists of the three best performing augmentation approaches at cross-validation time. We perform a majority vote to predict the final label and report the result in Table~\ref{table:sota_comp}.

\subsection{Experiment Setup}

All models are trained on one NVIDIA Tesla K80, 16\,GB RAM; all code is written in Pytorch 1.10.0+cu111. 
For all models, we report the accuracy as well as the sensitivity. Given the nature of the task, we believe it is important to put emphasis on the ability of models to detect a case of AD rather than just the accuracy. We perform 10-fold and 5-fold cross validation for text and audio, respectively, and report the average accuracy across 5 different seeds. Implementation details are available in our repository.
We further evaluate the augmentation approaches in terms of label preservation using a model trained only on the original data. We then measure the accuracy for both classes. We also compute the KL-divergence on extracted features and the Mel-spectrogram distortion between the original and augmented samples for audio. For text, we compute the Levenshtein distance, semantic distance, and type token ratio (TTR). For semantics, we extract sentence embeddings with the \textit{Huggingface} model \textit{all-mpnet-base-v2}\footnote{https://huggingface.co/sentence-transformers/all-mpnet-base-v2} and compute the cosine distance.

\subsection{Results}
\label{sec:results}

The results of our experiments evaluated on the ADReSS dataset are displayed in Tables \ref{table:acc_text} and \ref{table:acc_audio} for text and audio, respectively, and a comparison to the challenge baseline as well as state-of-the-art results can be found in Table \ref{table:sota_comp}. We can see that while DNN-based approaches to augmentation are an effective way to improve model performance, some traditional and less computationally expensive methods work comparably well.
For the text augmentation, paraphrasing methods showed the best results, whether it is by \textit{Back-translation} (BT), which achieves the highest score of 84\% for Russian (RU) and 85\% for German (DE) for BERT and SVM respectively, or by using the Pegasus model. However, simpler noise injection methods such as \textit{EDA} and \textit{Mixup} are not far behind in terms of the performance increase. Generally, the augmentations reduced overfitting for the BERT Model  while the SVM tends to overfit for almost all techniques. However, we can observe that the introduction of new samples improved both models' generalisation abilities via \textit{Text generation} for text and with \textit{FragmentVC} for audio, with second best results for AST (73\%)  and best for RF (69\%).

Similarly as for text, while individual traditional methods only offered improvement in some cases for the audio domain, the approach of \textit{Random} ensemble performed just as well as a more complex DNN-based method of voice conversion. We can also see that the results of our experiments are often 

considerably 
higher for the test set than for cross-validation. While this might partly be simply due to the small size of the dataset we work with, it also shows the role of data augmentation as a regularisation method. We also note the fusion of the top-3 best models in Table~\ref{table:sota_comp} and achieve the best results with 86\%.\\ 
We also evaluate the techniques on the DementiaBank, but we did not notice considerable improvement on the model's performance. We attribute this to the more noisy nature of the dataset and its larger size.

\begin{table}[!h]
\centering
\resizebox{\columnwidth}{!}{%
\begin{tabular}{lllllll}
\toprule

{} & \multicolumn{2}{c}{\textbf{BERT}} & \multicolumn{2}{c}{\textbf{SVM}} & \multicolumn{1}{c}{\textbf{BERT}} & \multicolumn{1}{c}{\textbf{SVM}}  \\
\cmidrule(lr){2-3}
\cmidrule(lr){4-5}
\cmidrule(lr){6-7}
\textbf{Method} &           CV & Test & CV & Test & \multicolumn{2}{c}{Test Sensitivity} \\
\cmidrule(lr){1-5}
\cmidrule(lr){6-7}
\textbf{None}              &  .80 (.04) &  .78 (.06) & .83 (.00) & .81 (.00) & .81(.04) & .71 (.00)\\
\cmidrule(lr){1-5}
\cmidrule(lr){6-7}
\rule{0pt}{\normalbaselineskip}\textbf{SD}                &  .74 (.01) &  .77 (.02) & .83 (.00) & .81 (.00) & .68 (.06) & .75 (.00) \\
\textbf{Mixup}             &  .79 (.04) &  .82 (.04) & .85 (.00) & .75 (.00) & .79 (.06) & .67 (.00) \\
\textbf{EDA }              &  .80 (.02) &  .83 (.03) & .84 (.00) & .79 (.00) & .81 (.04) & .71 (.00) \\
\textbf{Paraphrasing}        &  .82 (.05) &  .81 (.03) & .84 (.00) & .81 (.00) & .78 (.04) & .75 (.00) \\
\textbf{Text Gen. }  &  .71 (.02) &  .81 (.05) & .77 (.00) & .81 (.00) & \textbf{.84 (.07)} & .75 (.00) \\
\textbf{Context Aug.} &  .80 (.05) &  .80 (.05) & .83 (.00) & .77 (.00) & .80 (.06) & .67 (.00) \\
\textbf{BT (DE)}           &  \textbf{.82 (.02)} &  \textbf{.84 (.02) }& .\textbf{86 (.00)} & .81 (.00) & .81 (.06) & .75 (.00)\\
\textbf{BT (RU)}           &  .80 (.03) &   .82 (.01) & \textbf{.86 (.00)} & \textbf{.85 (.00)} & .83 (.06) & \textbf{.79 (.00)} \\
\bottomrule
\end{tabular}%
}
\caption{Accuracy and test Sensitivity results for Text augmentations on the ADReSS set, with standard deviation in parentheses. For the SVM model, once parameters are fixed, changing the random seed does not affect results, hence the standard deviation is zero. SD = random sentence deletion, BT = back-translation from either German (DE) or Russian (RU). Best results for each model are highlighted.}
\label{table:acc_text}
\end{table}

\vspace{-0.3cm}
\begin{table}[!h]
\centering
\resizebox{\columnwidth}{!}{%
\begin{tabular}{lllllll}
\toprule
{} & \multicolumn{2}{c}{\textbf{AST}} & \multicolumn{2}{c}{\textbf{RF}} & \multicolumn{1}{c}{\textbf{AST}} & \multicolumn{1}{c}{\textbf{RF}} \\
\cmidrule(lr){2-3}
\cmidrule(lr){4-5}
\cmidrule(lr){6-7}
\textbf{Method} &           CV & Test & CV & Test & \multicolumn{2}{c}{Test Sensitivity} \\
\cmidrule(lr){1-5}
\cmidrule(lr){6-7}
\textbf{None}         &  .62 (.13) &  .70 (.03) &  .58 (.14) &  .67 (.02) & 1.00 (.00) & .79 (.00) \\
\cmidrule(lr){1-5}
\cmidrule(lr){6-7}
\rule{0pt}{\normalbaselineskip}\textbf{Random}        &  .63 (.09) &  \textbf{.74 (.03)} &  .79 (.12) &  .68 (.02) & .92 (.08) & .79 (.00)\\
\textbf{Masking }      &  .56 (.11) &  .70 (.04) &  .69 (.16) &  .67 (.02) & .\textbf{98 (.02)} & .75 (.00) \\
\textbf{Loudness }     &  \textbf{.64 (.07)} &  .70 (.01) &  .78 (.14) &  \textbf{.69 (.02)} & .83 (.13) & .79 (.00) \\
\textbf{Noise  }       &  .62 (.17) &  .71 (.04) &  .69 (.15) &  .64 (.01) & .83 (.04) & .79 (.00) \\
\textbf{Pitch }        &  .61 (.13) &  .72 (.02) &  .68 (.15) &  .64 (.02) & .83 (.00) & \textbf{.88 (.00)} \\
\textbf{VTLP}          &  .61 (.07) &  .70 (.02) &  .79 (.13) &  .67 (.02) & .79 (.04) & .75 (.00) \\
\textbf{Shift }        &  .59 (.11) &  .72 (.02) &  .79 (.14) &  .66 (.02) & .79 (.04) & .75 (.00) \\
\textbf{Speed }        &  .63 (.07) &  .70 (.02) &  \textbf{.80 (.10)} &  .64 (.02) & .81 (.02) & .75 (.00)\\
\textbf{Normalisation} &  .56 (.07) &  .72 (.03) &  .76 (.11) &  .68 (.01) & .79 (.04) & .83 (.00)\\
\textbf{SpecAugment}   &  .56 (.10) &  .70 (.02) &  .53 (.07) &  .67 (.02) & .90 (.02) & .79 (.00)\\
\textbf{FragmentVC}    &  .62 (.04) &  .73 (.04) & .54 (.04) & \textbf{.69 (.03)} & .94 (.06) & \textbf{.88 (.00)}\\
\bottomrule
\end{tabular}
}
\caption{Accuracy and test Sensitivity results for Audio augmentations on the ADReSS set, with standard deviation in parentheses. The models used are a Random Forest (RF) and an Audio Spectrogram Transformer (AST).  VTLP = Vocal tract length perturbation. The best results for each model are highlighted.}
\label{table:acc_audio}
\end{table}
\vspace{-0.3cm}

\begin{table}[!ht]
\centering
\resizebox{\columnwidth}{!}{%
\begin{tabular}{llllllll}
\toprule

  &  \textbf{Model} & \textbf{Features} & \textbf{Accuracy} \\
\midrule
Baseline & LDA$^*$~\cite{luz2020alzheimer} & Text & .75 \\
 & LDA$^*$ (ComParE features)~\cite{luz2020alzheimer} & Audio & .63 \\
\midrule
SOTA & ERNIE (ensemble)~\cite{Yuan20} & Text & \textbf{.85} \\
 & Music-Linear-BOW~\cite{syed2021automated} & Audio & \textbf{.74} \\
 & Fusion~\cite{syed2021automated} & Audio + Text & \textbf{.90} \\
\midrule
Our models & SVM (BT RU) & Text & \textbf{.85} \\
 & AST (Random) & Audio & \textbf{.74} \\
 & Fusion: AST (Random) + SVM (BT RU) + BERT (BT DE) & Audio + Text & .86 \\

\bottomrule
\end{tabular}%
}
\caption{Comparison between our systems and SOTA on the ADReSS dataset. There are no baseline models or easily comparable work for the Dementiabank dataset. For our models, the model name and augmentation method used is provided. The best result is highlighted for each domain.\\ 
$^*$LDA = Linear discriminant analysis.}
\label{table:sota_comp}
\end{table}
\vspace{-0.3cm}

\label{sec:label_pres}
We note the label preservation performance as well as divergence from the original data distribution methods in Table \ref{table:label_pres_x1}. Methods with the highest label preservation scores tend hurt rather than benefit the model performance, suggesting that in those cases the augmented samples remain too similar to the original ones and the model ends up overfitting. At the same time, methods that diverge too much are likely to change key discriminative features of AD, also leading to poor performance. We can see that while noise injection and generative methods for audio have a considerably higher divergence score than the other traditional methods, they still have a beneficial effect on training. Comparing Table~\ref{table:acc_text} and Table~\ref{table:label_pres_x1}, we can see that although DNN-based approaches lower the label preservation, the performance on the test set and cross-validation sets indicate that the introduced noise decrease the models overfitting. In particular, \textit{Text generation} is the most disruptive in terms of all metrics but in fact improves on the accuracies and test sensitivity for both BERT (81\% \& 84\%) and SVM (81\% \& 75\%). \textit{Backtranslation} seems to strike the right balance between deviating from the training set (semantics, introduced new tokens) and preserving meaningful features, and similarly \textit{FragmentVC} for audio.

\begin{table}[!h]
\centering
\resizebox{\columnwidth}{!}{%
\begin{tabular}{llllccc}
\toprule
 & &  \multicolumn{2}{c}{Label Pres.} & \multicolumn{3}{c}{\textbf{$\Delta$}}\\
\cmidrule(lr){3-4}
\cmidrule(lr){5-7}
\textbf{Domain}  &   \textbf{Method}            & Acc. & F1 & Levenshtein & Semantics & TTR \\
\midrule
Text & \textbf{None} & .97  & 1.00 & 0 & .00 & .00\\
\cmidrule(lr){2-7}
      & \textbf{SD} &    .90 &   .90  & 71 & .08 & .03 \\
      & \textbf{Mixup} &    .92 &   .92 & 258 & .20 & .09\\
      & \textbf{EDA} &    .88 &   .88  &40 & .07& .01\\ 
      & \textbf{Paraphrasing} &     .79 &   .75  & 267 &.19 & .10\\
      & \textbf{Text Generation} &    .62 &   .69 &411 & .39 &.14\\
      & \textbf{Context Aug.}  &    .88 &    .89  & 48 &.09 & .03 \\
      & \textbf{BT (DE)} &    .83 &   .83  & 178 & .15 & .05\\
      & \textbf{BT (RU)} &    .83 &   .84  &228 & .21 & .03\\
      \midrule
      & & & &Divergence & Mel. Distortion \\
\midrule
Audio & \textbf{None} &1.00 &1.00 & .00 & .00 & -\\
\cmidrule(lr){2-7}
      & \textbf{Random} &    .90 &   .91 & .07 & 5.4 & -  \\
      & \textbf{Masking} &    .87 &   .90 & .07 & 11.7 & - \\
      & \textbf{Loudness} &    .95 &   .96 & .07 & 1.2 & - \\
      & \textbf{Noise} &    .84 &   .85 & .20 & 6.9 & -  \\
      & \textbf{Pitch shift} &    .80 &   .83 & .07 & 3.5 & - \\
      & \textbf{VTLP} &    .98 &   .98 &  .06 & 5.2 & - \\
      & \textbf{Time shift} &    .99 &   .99 &  .05 & 5.9 & -  \\
      & \textbf{Speed} &    .95 &   .96 & .06 & 5.1 & -   \\
      & \textbf{Normalisation} &    .88 &   .90 & .06 & 3.8 & -  \\
      & \textbf{SpecAugment} & .58  & .60   & .37 & 19.6 & -  \\
      & \textbf{FragmentVC}      & .58 & .52 & .39 & 9.7 & - \\
\bottomrule
\end{tabular}%
}
\caption{Label preservation and descriptors distances for approaches on Text and Audio domains on the  ADreSS set.}
\label{table:label_pres_x1}
\end{table}

\vspace{-0.6cm}

\section{Conclusion}
\label{sec:conclusion}

In this paper, we investigated a range of data augmentation techniques for the domains of text and speech. We evaluate the strategies on the ADReSS challenge dataset for the task of dementia detection for neural, SVM, and Random Forest models as well as their late fusion. Additionally, we quantify the label preservation properties for all augmentation approaches. Our experiments show that most of the techniques can help the models generalise and reduce overfitting on the training set. Generally we observe that neural-based augmentations do not necessarily outperform simpler approaches such as adding noise or performing random deletions and substitutions.
Our models achieve results comparable to state-of-the-art approaches on the ADReSS set by simply augmenting the training set.  Their optimal combination needs to be analysed next.

\bibliography{mybib.bib}
\bibliographystyle{IEEEtran}

\end{document}